\documentclass{article}
\usepackage{arxiv}
\usepackage{tcolorbox}
\usepackage{booktabs}
\usepackage{amsthm}
\usepackage{xcolor} 
\usepackage{framed} 
\usepackage{amsmath}
\usepackage{enumitem}
\usepackage{scrextend}
\usepackage[utf8]{inputenc} 
\usepackage[T1]{fontenc}    
\usepackage{hyperref}       
\usepackage{url}            
\usepackage{booktabs}       
\usepackage{amsfonts}       
\usepackage{nicefrac}       
\usepackage{microtype}      
\usepackage{lipsum}
\usepackage{graphicx}
\graphicspath{ {./images/} }
\usepackage{amsmath}
\usepackage{tgbonum}
\usepackage{titlesec}
\usepackage{algorithm}
\usepackage{fancybox}
\usepackage{algpseudocode}
\usepackage{geometry}  
\usepackage[utf8]{inputenc}  
\usepackage[T1]{fontenc}     

\newcounter{subsubsubsection}[subsubsection]
\renewcommand\thesubsubsubsection{\thesubsubsection.\arabic{subsubsubsection}}

\titleformat{\subsubsubsection}
  {\normalfont\normalsize\bfseries}{\thesubsubsubsection}{1em}{}
\titlespacing*{\subsubsubsection}
{0pt}{3.25ex plus 1ex minus .2ex}{1.5ex plus .2ex}

\makeatletter
\renewcommand\paragraph{\@startsection{paragraph}{5}{\z@}%
  {3.25ex \@plus1ex \@minus.2ex}%
  {-1em}%
  {\normalfont\normalsize\bfseries}}
\renewcommand\subparagraph{\@startsection{subparagraph}{6}{\parindent}%
  {3.25ex \@plus1ex \@minus .2ex}%
  {-1em}%
  {\normalfont\normalsize\bfseries}}
\def\toclevel@subsubsubsection{4}
\def\toclevel@paragraph{5}
\def\toclevel@subparagraph{6}
\def\l@subsubsubsection{\@dottedtocline{4}{7em}{4em}}
\def\l@paragraph{\@dottedtocline{5}{10em}{5em}}
\def\l@subparagraph{\@dottedtocline{6}{14em}{6em}}
\makeatother

\titleformat*{\section}{\LARGE\bfseries}
\titleformat*{\subsection}{\Large\bfseries}
\titleformat*{\subsubsection}{\large\bfseries}
\titleformat*{\paragraph}{\large\bfseries}

\title{LLMs Will Always Hallucinate, and We Need to Live With This}

\author{
 Sourav Banerjee* \\
  DataLabs\\
  United We Care\\
  \texttt{sb@unitedwecare.com} \\
   \And
Ayushi Agarwal\\
  DataLabs\\
  United We Care\\
  \texttt{ayushi@unitedwecare.com}
  \And
 Saloni Singla \\
  DataLabs\\
  United We Care\\
  \texttt{saloni@unitedwecare.com}
}

\begin{document}
\maketitle
\begin{abstract}
As Large Language Models become more ubiquitous across domains, it becomes important to examine their inherent limitations critically. This work argues that hallucinations in language models are not just occasional errors but an inevitable feature of these systems. We demonstrate that hallucinations stem from the fundamental mathematical and logical structure of LLMs. It is, therefore, impossible to eliminate them through architectural improvements, dataset enhancements, or fact-checking mechanisms. Our analysis draws on computational theory and Gödel's First Incompleteness Theorem, which references the undecidability of problems like the Halting, Emptiness, and Acceptance Problems. We demonstrate that every stage of the LLM process—from training data compilation to fact retrieval, intent classification, and text generation—will have a non-zero probability of producing hallucinations. This work introduces the concept of "Structural Hallucinations" as an intrinsic nature of these systems. By establishing the mathematical certainty of hallucinations, we challenge the prevailing notion that they can be fully mitigated. 

\end{abstract}


\section{Introduction}
\subsection{Background}
Large Language Models (LLMs) have achieved remarkable progress, demonstrating high fluency and accuracy in tasks like translation and question answering. They are increasingly influential in fields such as healthcare and education \cite{3, 4}. Future models are expected to be even more contextually aware, efficient, and ethically aligned. However, fundamental questions persist[5]: Can these models truly understand language without experiencing the world, and what are the risks of relying on artificial constructs for tasks traditionally requiring human judgment? A \textit{key} concern is hallucinations—instances where models generate plausible but incorrect information \cite{3,4,5}. Despite advances, these errors remain an intrinsic challenge, raising the question: Can we ever fully eliminate hallucinations from LLMs, or is this a limitation we must learn to manage?

\subsubsection{The Essence of Large Language Models}
At the core of large language models lies a deceptively simple principle: the prediction of linguistic patterns. \cite{5, 6}. The fundamental operation of an LLM can be distilled to a single, powerful question: 

\textit{Given a sequence of words, what word is likely to come next?} 

Assuming we refer to a transformer-like architecture \cite{72}, for a sequence of tokens $x = (x_1, x_2, ..., x_n) \text{where} x_i \in V$, a language model computes \cite{15}:

\begin{equation*}
  P(x) = P(x_1, x_2, ... , x_n) = \prod_{i=1}^nP(x_i| x_1, ... , x_{i-1})
\end{equation*}

where the conditional probability of the next token given the previous tokens is defined in the usual way \cite{16}:

\begin{equation*}
    P(x_{n+1}| x_1, ..., x_n) = \frac{P(x_1, x_2, ... x_n, x_{n+1})}{P(x_1, x_2, ... x_n)} 
\end{equation*}

As we scale these models, we observe a fascinating phenomenon: the emergence of apparently intelligent behaviours. \cite{46} Impressive as they are, the model has not learned to think and has no concept of truth; it has learned to mimic the products of thought with astonishing fidelity.

\subsection{Architectures of Large Language Model Generation}

\subsubsection{How language model generations work}

At the most basic level, LLMs work with tokens \cite{44} the fundamental units of text processing \cite{15}. These can be words, parts of words, or even individual characters \cite{15}. Special tokens play crucial roles in the model's operation \cite{46}. Let us consider a few \cite{46}: 

{\fontfamily{qcr}\selectfont
<BOS>: Beginning of Sequence \newline
<EOS>: End of Sequence 
}

Imagine we are feeding this sentence into our model: 

{\fontfamily{qcr}\selectfont
<BOS> The cat sat on the mat. <EOS>
}

The model first converts each token into a numerical representation - an embedding \cite{46,15}. But these embeddings alone lack positional encoding \cite{46}.
Absolute positional encoding \cite{47} addresses this by adding a unique pattern to each token's embedding based on its position, using sine and cosine functions \cite{47} allowing the model to distinguish the word order. This is done through roughly the following steps \cite{49}:

\begin{enumerate}
  \item Create a vector embedding as a function of semantics
  \item Create a separate positional encoding vector
  \item Add these two vectors to get a sum that stores both meaning and position
\end{enumerate}
 
This is the formulation for absolute positional encoding. For our sentence, it may look like this: 

\begin{tcolorbox}[colback=white, colframe=black, width=\textwidth, arc=0mm, auto outer arc]
\textbf{Position Encodings:} \\
Position 0 (BOS): [0.0000, 1.0000, 0.0000, 1.0000, \dots] \\
Position 1 (The): [0.8415, 0.5403, 0.0100, 0.9999, \dots] \\
Position 2 (cat): [0.9093, -0.4161, 0.0200, 0.9998, \dots] \\
\dots
\end{tcolorbox}

The self-attention \cite{47}  mechanism then determines how much each word "attends" to others, helping the model capture relationships like "the mat." 
The usual self-attention mechanism goes somewhat like the following \cite{49}: 

There are three tokens: \textit{query}, \textit{key}, and \textit{value}. 
\begin{itemize}
    \item The \textit{\textit{key}} vector is a representation of  the token whose output we want, $k_n=f_k(x_n, n)$ 
    \item The \textit{\textit{query}} vector is a representation of the token with which we want to check the relationship of the \textit{\textit{key}}: $q_m=f_q(x_m, m)$
    \item The \textit{\textit{value}} is another representation of the token whose output we want: $v_n=f_v(x_n, n)$
\end{itemize}

\textit{Attention} is calculated as a softmax over the dot products of the \textit{query} and \textit{key}. In essence, the dot product can be thought of as a distance measure between the \textit{query} and \textit{key}: \\
\begin{equation*}
a_{m,n} = \frac{exp(\vec{q_m }\cdot \vec{k_n})}{\Sigma_iexp(\vec{q_i}\cdot \vec{k_n)}}
\end{equation*}

The final output is then a weighted sum of the \textit{value} vector: 

\begin{equation*}
    o_n = \Sigma_ia_{i,n}v_n
\end{equation*}

For our sentence, it might compute something like the following table:

\begin{table}[h!]
\centering
\centering
\begin{tabular}{l|c|c|c|c|c|c}
    & \textbf{The} & \textbf{cat} & \textbf{sat} & \textbf{on} & \textbf{the} & \textbf{mat} \\ \hline
\textbf{The} & 0.1 & 0.3 & 0.1 & 0.1 & 0.2 & 0.1 \\ \hline
\textbf{cat} & 0.2 & 0.2 & 0.3 & 0.1 & 0.1 & 0.1 \\ \hline
\textbf{sat} & 0.1 & 0.2 & 0.2 & 0.2 & 0.2 & 0.1 \\ \hline
\textbf{on}  & 0.1 & 0.1 & 0.2 & 0.2 & 0.2 & 0.2 \\ \hline
\textbf{the} & 0.3 & 0.1 & 0.1 & 0.1 & 0.2 & 0.2 \\ \hline
\textbf{mat} & 0.2 & 0.1 & 0.1 & 0.2 & 0.3 & 0.4 \\ 
\end{tabular}

\vspace{0.5cm} 

\caption{Transition table for the sentence "The cat sat on the mat". Each entry represents the probability of transitioning from the row word to the column word.}
\end{table}

However, absolute positional encoding has limitations, especially for long sequences. Relative positional encoding \cite{47} considers the distance between words instead of assigning fixed positions.

In our example: 
"cat" to "sat": distance = 1
"cat" to "mat": distance = 4

\subsubsubsection{Rotatory Positonal Encoding (RoPE)}
RoPE embeds both, absolute and relative positional encoding into a single vector \cite{49}. It draws from the properties of complex representation of numbers in the Argand plane and instead stores position using a multiplicative approach. 

Referring to the definitions of \textit{\textit{query}}, \textit{\textit{key}}, and \textit{\textit{value}} above, the functions $f_q, f_k, f_v$ are encoded as: 

\begin{equation*}
    f_q(\vec{x_m}, m) = (W_q\vec{x_m})e^{im\theta}
\end{equation*}
\begin{equation*}
    f_k(\vec{x_n}, n) = (W_k\vec{x_n})e^{in\theta}
\end{equation*}

This allows for a definition of an inner product g that stores relative position information since it depends only on the quantity $(m-n)$: 

\begin{equation*}
    g(\vec{x_m}, \vec{x_n}, m-n) = Re[(W_q\vec{x_m})(W_k\vec{x_n})*e^{i(m-n)\theta}]
\end{equation*}

where “Re” represents the real part of the complex number and “*” represents the complex conjugation operation. $\theta \in \mathbf{R}$ is a preset non-zero constant. This can be presented as a matrix multiplication, with a rotation matrix as a factor: 

\begin{equation*}
\displaystyle f_{\{q, k\}}(\vec{x_m}, m) = \begin{bmatrix} cos(m\theta)&-sin(m\theta) \\
sin(m\theta)&cos(m\theta) \end{bmatrix} \begin{bmatrix} W^{11}_{q, k} & W^{12}_{q,k} \\
W^{21}_{q,k} & W^{22}_{q,k}\end{bmatrix} \begin{bmatrix} x^{(1)}_m \\
 x^{(2)}_m \end{bmatrix}
 \end{equation*}

where the first factor is of the form of a rotation matrix. \\

$g$ can similarly be written using a rotation matrix. 

Specifically, incorporating the relative position embedding is straightforward: simply rotate the affine-transformed word embedding vector by amount of angle multiples of its position index \cite{49}. This can be generalised to vectors with more than 2 dimensions, by dividing a $d$ dimensional space into $d/2$ two-dimensional spaces. 

Beyond transformers and self-attention, alternative approaches are being explored to improve the efficiency of LLMs, addressing computational challenges and enhancing performance.

\subsubsection{Linear RNNs}
Linear Recurrent Neural Networks (LRNNs) combine the sequential processing of traditional RNNs with the parallelization benefits of convolutions, offering improved inference speed \cite{50}. LRNNs learn not only the parameters but the architecture itself.
LRNNs are unique in that they do not employ iterative methods like backpropagation. LRNNs have the following properties: 

\begin{enumerate}[label=1.2.2.\arabic*, leftmargin=4.5em, labelsep=1em, itemsep=0.5em]
    \item They have three main kinds of weights:
    \begin{itemize}
        \item Input to reservoir
        \item Weights for within the reservoir
        \item Reservoir to output
    \end{itemize}
    \item The output neurons need not necessarily be distinct from the input neurons.
    \item All the above-mentioned weights are compiled into a transition matrix:
    \[
    \begin{bmatrix}
    W_{out} \\
    W_{in} & W_{res}
    \end{bmatrix}
    \]
    \item This matrix is decomposed into the Jordan form:
    \[
    W = sJs^{-1}
    \]
    \item The eigenvalues of the Jordan matrix are then analyzed; blocks with the smallest eigenvalues are removed.
\end{enumerate}

This allows for a smaller network, with the reduction achieved in a single step. \\

\subsubsection{Mamba (Linear-Time Sequence Modeling with Selective State Spaces)}

Mamba\cite{51} marked a significant advancement in sequence modelling, offering an alternative to attention-based models like transformers. Built on the foundation of State Space Models (SSMs) it introduces a crucial innovation: selectivity. Mamba introduces data-dependent gating and adaptability in crucial matrices and discretization step size, enhancing performance. \\
\begin{equation*}
   \frac{dx}{dy} = A(\Delta)x(t)+ B(\Delta)u(t) 
\end{equation*}
\begin{equation*}
    y(t)= C(\Delta)x(t)+ D(\Delta)u(t)
\end{equation*}

where $\Delta=\Delta(u)$ is a data-dependent step size and $A(\Delta),B(\Delta),C(\Delta),D(\Delta)$ are data-dependent matrices allowing for flexible and efficient processing of sequential data. Mamba promises similar performance (and crucially similar scaling laws) to the Transformer while being feasible at long sequence lengths (say 1 million tokens) but with linear-time complexity. To achieve this long context, the Mamba authors remove the \textit{quadratic bottleneck} in the Attention Mechanism, enabling it to run up to 5 times faster than transformers. 

\begin{figure}[h!]
    \centering
    \includegraphics[width = 0.8\linewidth]{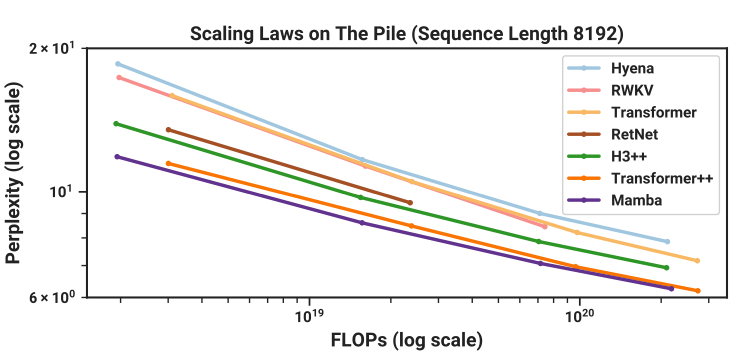} 
    \caption{Performance Comparison of Mamba and Other Language Models on The Pile Benchmark Dataset. Mamba exhibits comparable or slightly superior performance to other language models across various metrics on The Pile, a comprehensive dataset designed to evaluate the generalization capabilities of language models. Data reproduced from \cite{55}.}
\end{figure}

Following the Mamba model, the \emph{Jamba} \cite{73} hybrid model integrates Transformer and Mamba layers with a mixture-of-experts (MoE) module. It leverages the parallel processing strengths of Transformers and the efficiency of state-space models (SSMs) like Mamba by using data-dependent matrices and step sizes. The state-space equations are similar to the ones used for Mamba. Jamba can handle long contexts up to 256K tokens while maintaining a KV cache memory requirement that is significantly smaller than that of comparable models and achieves throughput up to 3x higher than similar models, combining powerful performance with remarkable resource efficiency.\cite{73}

\subsubsection{Kolmogorov-Arnold Networks (KANs)}

Traditional neural networks composed of multilayer perceptrons are based on the Universal Representation Theorem, which states that “any continuous function $f : [0, 1]^n \rightarrow [0, 1]$ can be approximated arbitrarily well by a neural network with at least 1 hidden layer with a finite number of weights.” \cite{60}. KANs \cite{53} draw on the Kolmogorov-Arnold representation theorem. They approximate complex multivariate functions through compositions of simpler, single-variable functions. The Kolmogorov-Arnold representation theorem therefore states that any multivariate function $f$ on a bounded domain can be represented as:

\begin{equation*}
    f(x_1, ... , x_n) = \sum\limits_{q=0}^{2n}\Phi_q(\sum\limits^n_{p=1}\Psi_{p,q}(x_p))
\end{equation*}
where:\\
$\Phi_q: \mathbf{R}\rightarrow \mathbf{R}$ and 
$\Psi_{p,q}: [0,1] \rightarrow \mathbf{R}$ \\

are continuous single-variable functions. Essentially, this means that the only “true multivariate function is addition since any other function can be represented by a composition of univariate functions and the sum”.\cite{59} The main difference between KANs and conventional multilayer perceptrons is that KANs employ learnable activation functions on edges, instead of the fixed activation functions on nodes. 
\begin{figure}
    \centering
    \includegraphics[width = 0.8\linewidth]{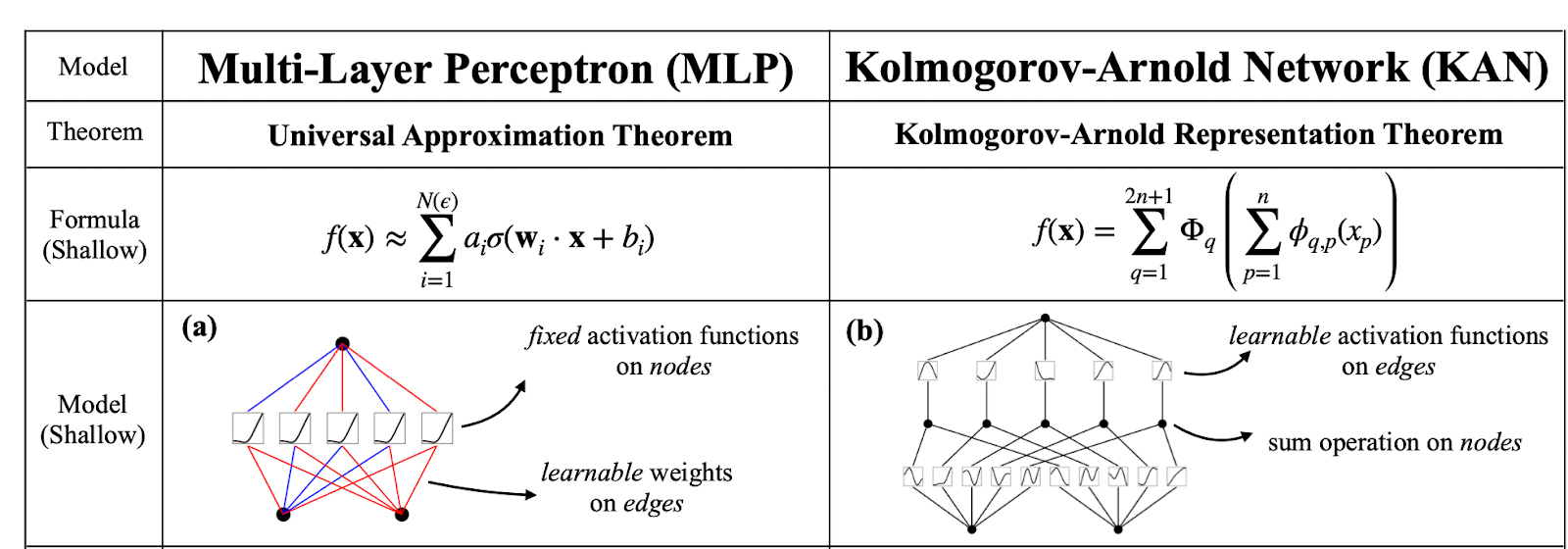}
    \caption{A comparison between multilayer perceptrons and KANs. Reproduced from \cite{53}.}
    \label{fig:3}
\end{figure}
\subsection{Transfer Learning}
Despite the architectural diversity, a fundamental truth persists: LLM generation is prone to hallucination. There exist complementary techniques that do not alter core architectures but still significantly impact the performance of LLMs, which often tend to be extremely bulky with billions of parameters. \cite{61,62}. Pre-trained models \cite{61}, though versatile, are not always perfectly suited for every application 
\cite{63}. Transfer learning techniques train the model's parameters to better align with specific data, improving the model's performance \cite{65}.

\subsubsection{Parameter-Efficient Fine-Tuning (PEFT): Streamlined Approaches}
Traditional fine-tuning updates all model parameters, which can be computationally expensive and resource-intensive, especially for large models[64,65,66]. Parameter-Efficient Fine-Tuning (PEFT)\cite{64} updates a much smaller number of parameters, reducing computational costs, but still helping LLMs adapt to specific tasks. These smaller numbers of parameters may be new additions, or a selected subset of the existing parameters, or a hybrid of both \cite{54}. Below are some \textit{key} approaches.

\subsubsubsection{Adapters}
were among the first methods to use a smaller number of parameters for fine-tuning \cite{64}. They introduce additional, small trainable modules into the model, updating only these during fine-tuning \cite{64,65}. This helps reduce computation costs while maintaining the accuracy provided by vanilla fine-tuning. \cite{65}

\begin{figure}[h!]
    \centering
    \includegraphics[width=0.7\linewidth]{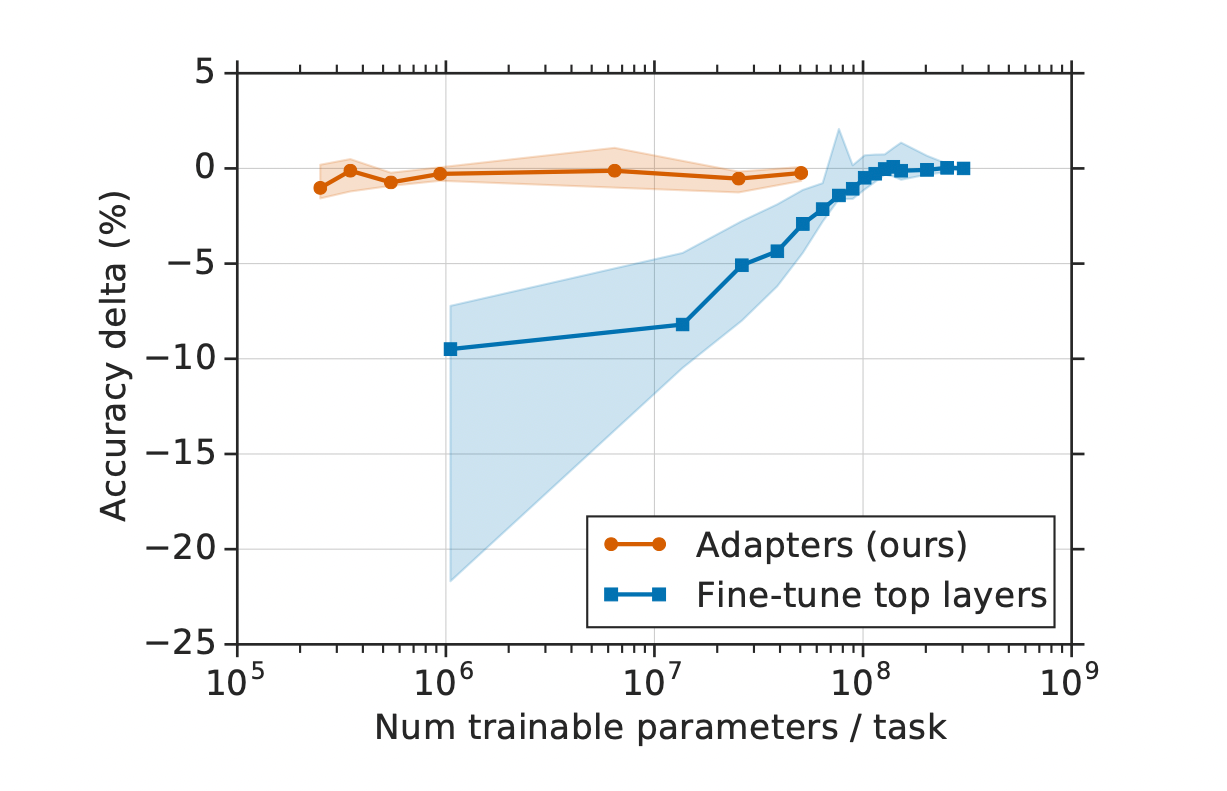}
    \caption{Improvement in accuracy as a function of the number of trainable parameters. Reproduced from \cite{65}.}
    \label{fig:4}
\end{figure}

\subsubsubsection{BitFit} (Bias Terms Fine-Tuning) modifies only the bias terms b and the task-specific classification layer in the model[64,66]. This leads to a substantial reduction in the number of parameters used, as compared to vanilla fine-tuning, since it uses just a subset of the additive bias terms. \cite{66}

\subsubsubsection{Low-Rank Adaptation (LoRA)}
allows us to train some dense layers in a neural network indirectly by optimizing rank decomposition matrices of the dense layers' change during adaptation instead, while keeping the pre-trained weights frozen. \cite{67} 

Let us represent the updated weight matrix as $W'$. Then, 
\begin{equation*}
    W' = W_0 + {\Delta}W
\end{equation*}
Where $W_0$ is the pretrained weight matrix and $W$ is the accumulated gradient update. 

LoRA reduces the dimensionality of ${\Delta}W$ by factorising it into low-rank matrices $A$ and $B$:
\begin{equation*}
    {\Delta}W = BA
\end{equation*}

The beauty of this procedure lies in a simple observation from linear algebra: Two lower dimensional matrices can be multiplied to create a higher dimensional matrix. 

Let us assume that $W_0$ is of the dimension $d{\times}m.$ Then, we may use two matrices $A\in \mathbf{R}^{d{\times}r}$ and $B \in \mathbf{R}^{r{\times}m}$, where we set $r << min(d, m)$ \cite{67}.
Hence, the number of parameters reduces significantly from $d \times m$ to $(d \times r) + (r \times m)$.

\subsubsection{Combining PEFT Techniques}
PEFT methods can be combined to further enhance efficiency. For example, one might use adapters in combination with LoRA \cite{61}. 

\subsubsection{Retrieval-Augmented Generation (RAG): Bridging Knowledge Gaps}

Fine-tuning improves how large language models (LLMs) perform on specific tasks, but it does not always ensure that the information they produce is factually accurate. Retrieval-Augmented Generation (RAG) addresses this by combining the strengths of language models with information retrieval systems, allowing the model to generate content based on accurate, up-to-date information.

In simple terms, RAG works like this: 

\begin{equation*}
    y = G(x, R(x))
\end{equation*}

Here, $y$ is the output, $G$ is the language model, $x$ is the input, and $R(x)$ is the relevant information retrieved from an external knowledge base.

\textbf{The Retrieval Process}\\

\emph{Vector Representations}: In RAG, both the input query and the documents in the knowledge base are turned into dense vectors:
\begin{equation*}
   e_q = E_q(q) 
\end{equation*}
\begin{equation*}
   e_d = E_d(d) 
\end{equation*}

These vectors, created by encoding functions  $Eq$ and $Ed$, represent the meaning of the text in a way that can be easily compared.

\emph{Similarity Measures}: To find the most relevant documents, RAG compares the vector of the query $e_q$ with the vectors of documents $e_d$ using similarity measures like: 
\begin{itemize}
    \item Cosine Similarity checks how similar the directions of two vectors are, with results ranging from -1 (very different) to 1 (very similar)
    \begin{equation*}
        cos(e_q, e_d) = \frac{e_q \cdot e_d}{||e_q|| ||e_d||}
    \end{equation*}

  \item 
  Dot Product multiplies the components of the vectors, providing a straightforward way to compare them 
  \begin{equation*}
      s(e_q, e_d) = e_q \cdot e_d
  \end{equation*}
\end{itemize}

\textbf{Retrieval and Ranking}\\
\\ 
After calculating similarity scores, the system retrieves the top $k$ most relevant documents. This involves two main steps:
\begin{itemize}
    \item Approximate Nearest Neighbor (ANN) Search: To quickly find the best matches from a large knowledge base, ANN algorithms like Hierarchical Navigable Small World (HNSW) are used.

    \item Re-rank these top $k$ documents using more detailed methods to ensure the most accurate information is selected.

\end{itemize}

\textbf{The Generation Process} \\

\emph{Augmentation}: Once the most relevant documents are retrieved, they are combined with the original input: 
\begin{equation*}
    x_{augmented} = [x:R(x)]
\end{equation*}

This augmented input gives the language model more context, grounded in factual information

\emph{Output Generation}: The language model then generates the final output based on this augmented input:
\begin{equation*}
    y = G(x_{augmented})
\end{equation*}

This process helps ensure that the model's output is not only relevant but also based on accurate information.
\subsection{Hallucinations in LLMs: What They Are and How They Happen}

Hallucinations in large language models (LLMs) occur when the models generate content that is false, fabricated, or inconsistent with their training data. These happen when the model, in an attempt to produce coherent responses, fills in gaps with plausible-sounding but incorrect information. Hallucinations can range from subtle inaccuracies to completely fictional assertions, often presented with high confidence. It is important to note that LLM hallucinations can occur even with the best training, fine-tuning, or the use of techniques like Retrieval-Augmented Generation (RAG). 

\subsubsection{Types of Hallucinations in LLMs}
We have identified four main types of hallucinations that can occur in large language models (LLMs):

\subsubsubsection{\textbf{Factual Incorrectness: When AI Gets It Wrong}}

Factual inaccuracies occur when LLMs provide incorrect information based on existing data, but without inventing new, non-existent details. For example, an LLM might incorrectly state a patient's blood sugar level as 150 mg/dL when the correct \textit{value} is 120 mg/dL. This type of error arises from mishandling of factual data within the model's knowledge base. These inaccuracies can be dangerous in contexts where precision is critical, such as healthcare, where an incorrect \textit{value} could lead to inappropriate treatment.

\subsubsubsection{\textbf{Misinterpretation}}

Misinterpretation occurs when LLMs fail to correctly understand input or context, leading to inaccurate responses. There are two primary forms of misinterpretation: \\

\emph{Corpus Misinterpretation}: The model misclassifies the intent or context within its vast knowledge base, resulting in a response that does not accurately reflect the intended meaning.\\
\\
\emph{Prompt Misinterpretation}: The model misinterprets the user's input due to ambiguous wording or its own limitations. For example, the question "What is the meaning of lead?" might be wrongly interpreted as a \textit{query} about the chemical element instead of leadership, depending on the context.

\subsubsubsection{\textbf{Needle in a Haystack}}

The "Needle in a Haystack" problem refers to the challenge LLMs face in retrieving specific, correct information from a vast corpus. This can manifest in two ways: \\

\emph{Missed key Data Points:} The model may provide incomplete information, such as citing only one cause of World War I while omitting others. Mathematically, this occurs when the retrieval function R(x) fails to fully capture the necessary data: 
\begin{equation*}
    y = G(x, R(x_{partial}))
\end{equation*}

where $R(x_{partial})$ represents an incomplete set of retrieved information.\\

\emph{Partial Incorrectness: }The model might mix accurate facts with errors, such as incorrectly stating that Neil Armstrong walked on the moon in 1959 instead of 1969. This blending of correct $R(x_{correct})$ and incorrect $R(x_{incorrect})$ information results in outputs that are neither entirely true nor false: 
\begin{equation*}
    y = G(x, [R(x_{partial});R(x_{incorrect})]
\end{equation*}

\subsubsection{Fabrications}

Fabrications involve the creation of entirely false statements that have no basis in the model's training data. Unlike factual inaccuracies, where the model incorrectly represents existing data, fabrications are pure inventions by the model. For instance, an LLM might create a fictitious scientific study or invent a quote from a historical figure that never existed. Fabrications occur when the model generates output without any real supporting retrieval function: 
\begin{equation*}
    y = G(x, R(x_{fabricated}))]
\end{equation*}

where $R(x_{fabricated})=\phi$ indicates no actual retrieval, leading to the invention of entirely fictional content.

\subsubsection{Societal Consequences of LLM Hallucinations}
The societal consequences of hallucinations in LLMs can be far-reaching and serious:

\emph{Misinformation and Disinformation}: Hallucinated content can spread false information, leading to public misunderstanding on critical issues such as health, science, and politics. This can exacerbate the spread of misinformation and disinformation, undermining public trust in media and institutions. \\ \newline
\emph{Legal and Ethical Risks}: As seen in legal cases where LLMs fabricated information, hallucinations can lead to severe legal consequences, including wrongful convictions or legal disputes based on incorrect facts. \\ \newline
\emph{Impact on Public Health}: Inaccurate medical information generated by LLMs can lead to dangerous health outcomes if individuals or even healthcare providers rely on incorrect recommendations. \\ \newline
\emph{Erosion of Trust}: Repeated exposure to AI-generated hallucinations could erode public trust in AI systems, making people sceptical of even accurate AI-driven insights and solutions. \\ \newline
\emph{Amplification of Bias}: If hallucinations reflect or amplify existing biases in the data, they can contribute to social division and perpetuate harmful stereotypes.

\subsection{Identification of LLM Hallucination and Mitigation Strategies}

In this section, we look at some of the hallucination mitigation strategies that have been employed so far. 

\subsubsection{Chain-of-Thought (CoT) Prompting}
This section draws from the original paper\cite{48}. Researchers discovered that these models could perform complex reasoning tasks like “arithmetic, commonsense, and symbolic reasoning tasks” when prompted using Chain-of-Thought prompting. Chain-of-Thought (CoT) prompting encourages LLMs to make the reasoning process explicit and potentially reduce logical inconsistencies and hallucinations. 

For example:

\begin{tcolorbox}[colback=white, colframe=black, boxrule=0.5mm, width=\textwidth, arc=0mm, auto outer arc]
\textbf{Prompt}: "Q: Roger has 5 tennis balls. He buys 2 more cans of tennis balls. Each can has 3 tennis balls. How many tennis balls does he have now?"\\
\\
\textbf{Chain of Thought}: "Roger starts with 5 tennis balls. He buys 2 cans of tennis balls. Each can contains 3 tennis balls. So, from the cans, he gets: 2 * 3 = 6 tennis balls. Now, we add these to his original tennis balls: 5 + 6 = 11. Therefore, Roger now has 11 tennis balls."
\end{tcolorbox}

By making the reasoning process explicit, CoT helps reduce logical errors and nonsensical hallucinations. However, while CoT improved performance on many reasoning tasks, it did not eliminate hallucinations entirely. Models could still produce logically sound but factually incorrect reasoning chains.

\subsubsection{Self-Consistency}

Recognizing this, Wang et al \cite{58} proposed the concept of self-consistency. This was based on using CoT prompting to generate multiple reasoning paths and selecting the most consistent one. 

The process is as follows. The model is prompted with a CoT prompt. Now, assume that there are a set of answers ${a_1, a_2, ... }$  generated by the model for that prompt. Let us label the reasoning to get to an answer $a_i$ as $r_i$ . The model creates pairs $(r_i,a_i)$. Then, the model selects the most consistent answer as:

\begin{equation*}
    a = argmax_a(\sum\limits_{i=1}{m}\mathbf{1}(a=a_i))
\end{equation*}

This leverages the intuition that correct answers are more likely to be arrived at consistently, while hallucinations or errors are likely to be more varied. 

\subsubsection{Uncertainty Quantification}
This subsection relies on \cite{70} for information. To tackle the issue of models expressing high confidence even when hallucinating, researchers have introduced uncertainty quantification techniques\cite{68}.  Uncertainty quantification depends on the model used. 
\subsubsubsection{\emph{Softmax Neural Networks:}}  These are classifiers that predict the class based on the softmax function: 
\begin{equation*}
    p(y = z | x, \omega) = \frac{f^{\omega}_z(x)}{\sum\limits_{z'=1}^Mf^{\omega}_{z'}(x)}
\end{equation*}

“where $z$ is the given output class, which belongs to the set of all possible outcomes, $Z$. $X$ is the input sample and $f^{\omega}_{z'}(x)$ is an arbitrary function, parameterised by $\omega$, giving the support that $x$ belongs to class $z$”.

The class is given by $argmax_z(p(y = z | x, \omega))$. The probability distribution over the classes $z$ can help us quantify the uncertainty. 

Some studies also employ the gradient of the loss function with respect to the parameters, which is seen as a measure of the 'stability' of the model. 
\subsubsubsection{\emph{Bayesian Neural Network}}: This network treats the weights of the model as belonging to a distribution. The posterior distribution $\Omega(\omega| X, Y)$ of the model weights is found given the training data $X, Y$. We then take multiple samples of the weights  to create multiple models. Then, the final output class is given by: 
\begin{equation*}
    p(y | x, \Omega)= \frac{1}{M}\sum\limits_{i=0}^{M}f^{\omega_i}(x)
\end{equation*}

where $f^{\omega_i}$ is the neural network parameterized by the $i^{th}$  sample of the weights  from the posterior distribution. 

\subsubsubsection{\emph{Ensemble Neural Networks}} 
Much like a Bayesian Neural Network, this algorithm consists of a set of models as well. Here however, the models are independent of each other, and make their predictions separately from the other members in the set. Here the classification is given by: 
\begin{equation*}
    p(y | x, w_1, ... w_M)=\frac{1}{M}\sum\limits_{i=0}^{M}f^{w_i}(x)
\end{equation*}

where $M$ is the number of models, and each $w_i$ represents the parameters of an independent model. 

Uncertainty quantification itself can come many methods, some of which are discussed below. 
\subsubsubsection{\emph{Shannon entropy}} is defined as: 
\begin{equation*}
    H(y, X)=-\sum\limits_ip(y=i|X)log(p(y=i|X))
\end{equation*}

\subsubsubsection{\emph{Norm of the gradient of the loss with respect to the parameters}}: This is represented as $||\nabla_{\omega}L||$
where  $\nabla_{\omega}L=\nabla_{\omega}l(\hat{y}_i, f^{\omega}(x_i))$. However, uncertainty quantification only helps identify potential hallucinations -  it doesn't prevent them. Models can still be confidently wrong.

\subsubsection{Faithful Explanation Generation}
With LLMs used for critical applications, there is a need for explanations of how models arrive at conclusions. These explanations themselves must be evaluated. Faithfulness is one such parameter. It refers to the extent to which an explanation accurately reflects a model's reasoning process \cite{69}. This, in turn, helps users be aware of the generation process and, therefore, arms them to spot hallucinations.

One such measure of faithfulness is Shapley \textit{value}s \cite{71}. 

Shapley \textit{value}s, a game theory concept, fairly allocate payouts among contributors based on their input. In ML, features are the "players," and Shapley \textit{value}s measure each feature/data point's contribution to a prediction, helping users understand how the prediction was made.
Given an algorithm $A$, an arbitrary data set $D$, and an accuracy measure $V$, a data \textit{value} $\phi_i(D, A, V)$ quantifies the '\textit{value}' of the $i^{th}$ datum. To follow all the requirements of a fair distribution, this function should be of the following form: 
\begin{equation*}
    \phi_i= C \sum\limits_{S{\subseteq}D-{i}}\frac{V(S{\cup}i) - V(S)}{{{n-1}\choose{|S|}}} 
\end{equation*}

where $C$ is a multiplicative constant. $S$ is a subset of $D/i.$ 

\textbf{Conclusion}

Above, we have noted techniques that attempt to improve on every stage in the LLM output generation process and hence mitigate hallucinations. Let us look at each stage in turn.  \\

\emph{Training}: This is improved through architectural improvements, like transformers, KAN, Mamba, Jamba and others. Transformers introduced a self-attention mechanism to better maintain context across input sequences; KANs provide an alternative mathematical basis for neural networks, while Mamba and Jamba give an alternative to attention-based mechanisms. \\

\emph{Intent Classification}: This has been improved through various techniques like Chain-of-Thought prompting, RAG, among others. Chain-of-Thought was a major observation that allowed simple explanatory prompts to help the model perform better, while RAG is a method of information retrieval that complements any architecture.   \\

\emph{Information Retrieval}: This again can be improved through Chain-of-Thought prompting and RAG. These techniques help the model identify the correct pieces of information to be retrieved from its database.   \\

\emph{Output Generation}: Methods like Self-Consistency help the model select the best response at the output generation stage. \\

\emph{Post Generation Fact Checking}: Techniques like Uncertainty Quantification and Faithfulness Explanation Generation help users and models identify the correctness of generated responses. 
\begin{figure}[h!]
    \centering
    \includegraphics[scale = 0.3]{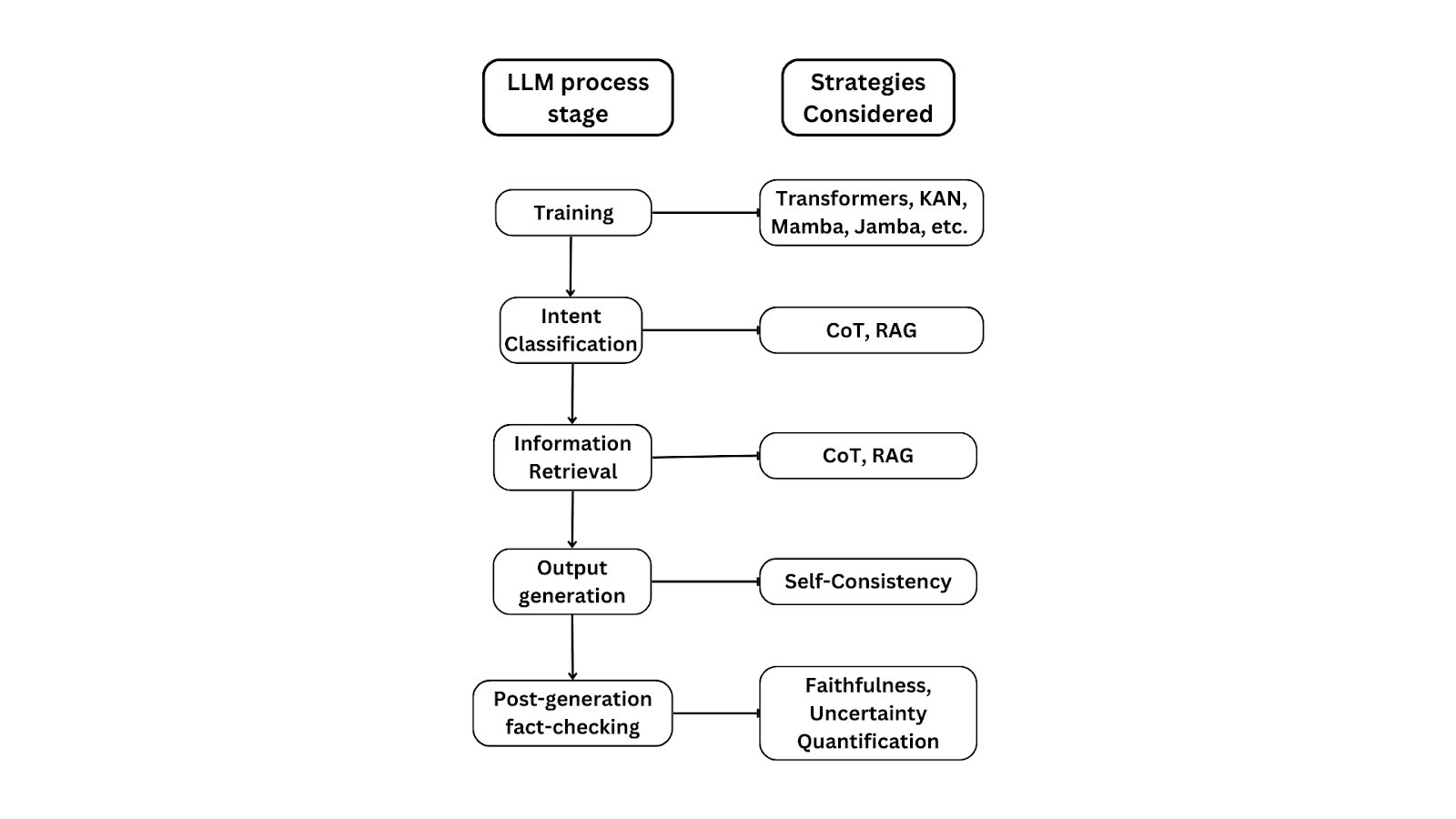}
    \caption{Stages of LLM Generation and Strategies to Mitigate Hallucination in Each of Them}
    \label{}
\end{figure}

Below, we will show that no matter which one these (or yet undiscovered) techniques one employs, the fact remains: LLMs will hallucinate; hallucinations can never be fully eliminated. \\

At every one of these stages, the LLMs are susceptible to hallucinations. Training can never be 100\% complete; intent classification and information retrieval are undecidable; output generation is necessarily susceptible to hallucination; and post-generation fact checking can never be a 100\% accurate: Irrespective of how advanced our architectures or training datasets, or fact-checking guardrails may be, hallucinations are \emph{ineliminable}. \\

Let us investigate each of these in turn in the coming sections. 

\section{All Hallucinations are Structural Hallucinations}

\subsection{Structural Hallucinations can never be eliminated from Large Language Models}

We introduce the concept of Structural hallucinations: they are an inherent part of the mathematical and logical structure of any LLM.

Consider language model output generation as a series of intricate steps—from the initial training to the final output. Each step carries a non-zero probability of a structural hallucination occurring regardless of the sophistication of our models or the vastness of our training data.

Let us examine this process more closely, unveiling the causes of hallucination at each critical stage:

\begin{enumerate}[label=2.1.\arabic*, leftmargin=4.5em, labelsep=1em, itemsep=0.5em, start=4]
    \item No training data can ever be complete. We can never give 100\% a priori knowledge. The vastness and ever-changing nature of human knowledge ensures that our training data will always be, to some degree, incomplete or outdated.
    \item Even if the data were complete, LLMs are unable to deterministically retrieve the correct information with 100\% accuracy. The very nature of these models ensures that there will always be some chance, however small, of retrieving incorrect or irrelevant information.
    \item An LLM will be unable to accurately classify with probability 1. There will always be some ambiguity, some potential for misinterpretation.
    \item No a priori training can deterministically and decidedly stop a language model from producing hallucinating statements that are factually incorrect. This is because:
    \begin{enumerate}[label=2.1.7.\arabic*, leftmargin=4em, labelsep=0.5em, itemsep=0.3em]
        \item LLMs cannot know where exactly they will stop generating. (LLM halting is undecidable - explained ahead)
        \item Consequently, they have the potential to generate any sequence of tokens.
        \item This unpredictability means they cannot know a priori what they will generate.
        \item As a result, LLMs can produce inconsistent or contradictory, as well as self-referential statements.
    \end{enumerate}
    \item We could attempt to fact-check, given a complete database. However, even if we attempt it, no amount of fact-checking can remove the hallucination with 100\% accuracy.
\end{enumerate}

\begin{figure}[h!]
    \centering
    \includegraphics[scale = 0.3]{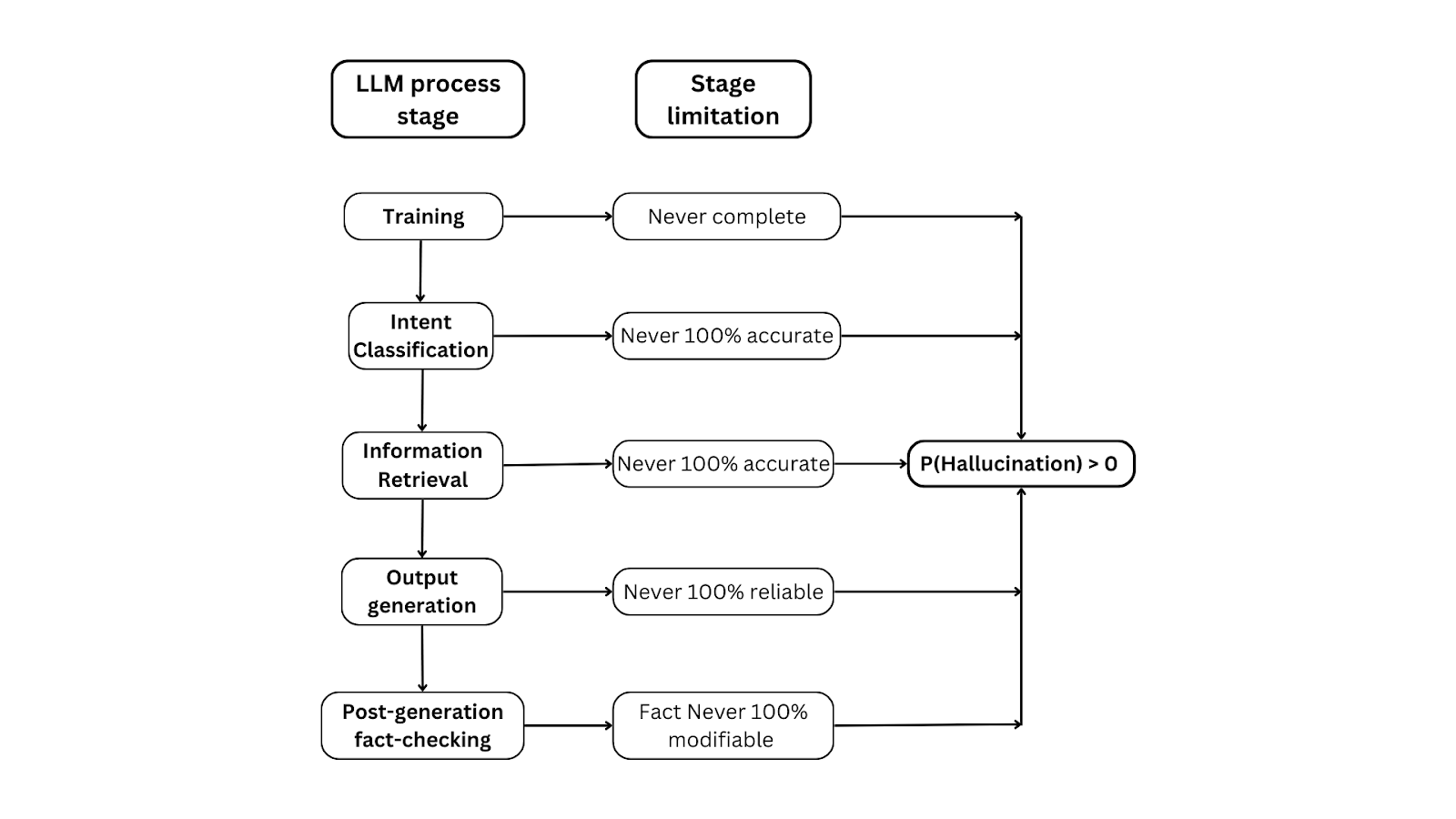}
    \caption{There are limitations associated with every stage of the LLM generation process. This leads to an inevitable non-zero probability of hallucination in LLMs}
    \label{}
\end{figure} 

Language models possess the potential to generate not just incorrect information but also self-contradictory or paradoxical statements. They may, in effect, hallucinate logical structures that have no basis in reality or even in their own training data. As we increase the complexity and capability of our models, we may reduce the frequency of these hallucinations, but we can never eliminate them entirely. 

\subsection{Some Preliminaries}

We explain below some of the concepts used in our proofs in section 3. 
This section is, for the most part, based on information from [1]. For further details, the reader is referred to \cite{1}.

\subsubsection{Turing Machines}
Turing Machines, conceptualised by Alan Turing in 1936, were developed to address Hilbert's question about determining mathematical truths mechanically. They consist of- 
\begin{itemize}
    \item A finite set of instructions that tell the machine what to do. 
    \item A tape divided into cells, on which the Turing Machine can write.
    \item The symbols used by the machine to write on the tape. They are collectively called the alphabet of the Turing Machine.
    \item A read-write tape head, with which the Turing Machine reads/writes.
    \item A state register, to keep track of the state. 
\end{itemize}

The machine computes by moving the tape head, reading/writing symbols, and changing states until reaching an accept or reject state. Turing Machines can simulate any algorithm, making them useful for modelling LLMs. A Universal Turing Machine can simulate any other Turing Machine.

\subsubsection{Decidability}
However, there are some very real limitations on the problems a computer can compute. No matter how hard we try or how many resources (like infinite memory) we dedicate to the computer, a computer cannot solve all problems. The idealisation of the Turing Machine helps us investigate computation's limitations, which are expressed in the language of undecidability. For example, consider the set of strings that have an equal number of 0s and 1s. We ask a “yes/no” question: Does a string x belong to this set? Our Turing Machine is able to give us a definitive answer. However, some “yes/no” questions cannot be answered definitively with an algorithm; there is no definitive set of steps to solve the given problem. These questions are undecidable by computers. 

In the following subsection, we will look at three problems that are undecidable on Turing Machines: The Acceptance problem and the Halting problem. 

\subsubsection{Some Undecidable Problems}
These are the undecidable problems to which we will repeatedly refer throughout the paper. We explain them here\cite{1}. Given a Turing Machine $M$ and an input string $w$:

\subsubsubsection{\emph{Halting problem}}: Does $M$ halt (i.e. either accept or reject, but not loop forever) on $w$?
\subsubsubsection{\emph{Acceptance Problem}}: Does $M$ accept $w$?

In essence, the undecidability of these problems means that an algorithm cannot predict its own behaviour—it cannot know a priori whether it will reject, accept, or run forever on an input.

\subsubsection{Reductions}
To show that problem A is undecidable, it is helpful to reduce a known undecidable problem B to the problem A at hand. This shows that A is also undecidable.

\textit{\textbf{Define:}}

\begin{itemize}
    \item $A$: The problem we're examining.
    \item $D_A$: A hypothetical decider for problem $A$.
    \item $B$: A known undecidable problem.
    \item $D_B$: A decider for problem $B$ that we construct using $D_A$.
\end{itemize}

\textit{\textbf{Proof:}}
\begin{itemize}
    \item Assume $A$ is decidable, i.e. $\exists D_A : D_A$ is a Turing Machine that decides $A$.
    \item Construct $D_B$ using $D_A$ as a subroutine.
    \item Given: $B$ is undecidable, i.e. $\neg\exists{D_B} : D_B$ is a Turing Machine that decides $B$.
    \item From (1) and (2): If $D_A$ exists, then $D_B$ exists.
    \item Contradiction: (3) contradicts (4), as the existence of $D_B$ implies $B$ is decidable.
    \item Therefore, our assumption in (1) must be false.
    \item Conclusion: $A$ is undecidable, $\neg\exists{D_A} : D_A$ is a Turing Machine that decides $A$.

\end{itemize} 



The second step above is what is called a reduction: We \emph{reduce} a known problem $B$ to our problem $A$. We call it a reduction because the problem 'reduces' to finding a way to solve $A$; in step (2), we show that if we solve $A$, then we can solve $B$ by building a decider for $B$ using the decider for $A$. 

\section{Hallucination is Inevitable: Claims and Proofs}

In section 2.1, we explored all the stages of LLM output generation and how they are susceptible to hallucination. In this section, we prove these claims.  

\subsection{Training Data is Inherently Incomplete}
\textbf{No training data can ever be complete; an LLM can never be provided with 100\% a priori knowledge}. 

All of an LLM's knowledge comes from data. Therefore, it seems to stand to reason that a larger, more complete dataset is a solution to hallucination; we should just give it all the knowledge in the world. Unfortunately, this is not possible. We demonstrate that while increasing the amount of data can tend towards a more complete database, it can never result in a 100\% complete database. Hence, a larger training set, no matter how large, cannot eliminate structural hallucinations. 

To this end, we claim the following: 
\begin{quote}
    \textbf{Assertion 1: No training dataset can contain all true facts.}
\end{quote}

In what follows, an arbitrary LLM itself will tell us that the training dataset cannot be 100\% complete. Our proof proceeds by contradiction. We begin by assuming there exists a training dataset $D$ that contains complete knowledge. This dataset $D$ is assumed to contain every possible fact about the world.

\subsubsection{Proof}

Let:
\begin{itemize}
    \item $D$ be the set of all facts in the training dataset.
    \item $F$ be the set of all true facts about the world.
\end{itemize}

\subsubsubsection{Considerations}
Consider now the statement $S_H$: $S_H$ = "It is a fact that there exist true facts beyond the facts in my training database."  This can be expressed as: \\
\begin{equation*}
    S_H = \exists x:x\in F \land x \notin D   
\end{equation*}
  
We analyse $S_H$ by cases: \\
\\
\textbf{Case 1: $S_H$ is False} \\
\begin{itemize}
    \item Let $G$ be the function representing the LLM's generation process. Then, since we only want the LLM to produce true statements, we ask every generation be a part of $F$, which is the set of all true facts:
\begin{equation*}
    G(D) \subseteq F
\end{equation*}
    \item By assumption, the LLM has generated $S_H$: 
\begin{equation*}
    S_H \in G(D)
\end{equation*}

\item But $S_H$ is false: 
\begin{equation*}
    S_H\in{G(D)}\land{S_H}\notin{F} 
\end{equation*}
($S_H$ is both a generation and false) 
\item This is in contradiction to step 3 above, which requires $G(D) \subseteq F$. With $S_H$, we have shown 
\begin{equation*}
    G(D) \not\subseteq F
\end{equation*}
which means that not all generations of the LLM are true, i.e. the LLM has hallucinated with the statement $S_H$.
\end{itemize}

\textbf{Case 2: $S_H$ is True}
\begin{itemize}
    \item Let us consider a set $X$:
\begin{equation*}
    X = {z : z{\in}F\land{z}\notin{D}}  
\end{equation*}

The set $X$ is the set of all true facts that are not in the model’s database $D$.

\item Then, 
\begin{equation*}
    	S_H{\in}X{\land}S_H{\in}G(D) 
\end{equation*}
	($S_H$ is a true fact that lies outside the database D and $S_H$ is generated by the LLM.)

\item Now, consider any statement $w$ present in $D$.\\
    $w\in{D}$ and,\\
    $\forall{w}{\in}D, w : w$ cannot verify $z{\in}X$.\\ 
    No fact in $D$ can verify $z$, which is a true fact that lies outside $D$. 
    
Then, $S_H{\in}G(D)$ cannot be verified by any $w{\in}D$.

\item Therefore, $G$ has produced a statement that cannot be verified using its training data. This unverifiable output constitutes a form of hallucination.
\end{itemize}

\subsubsubsection{Conclusion from Both Cases}
Let $H(G)$ denote "$G$ produces a hallucination".

From Case 1 and Case 2: \\
$({\neg}S_H\implies{H(G)}){\land}(S_H{\implies}H(G)) $

In both cases, whether $S_H$ is true or false, the LLM has hallucinated. 

This proves that regardless of the truth value of $S_H$, the LLM generation $G$ will produce a hallucination, demonstrating that:
\begin{equation*}
   \exists y : y{\in}G(D){\land}(y{\notin}F{\lor}(y{\in}F{\land}{\neg}({\exists}v{\in}D : v \text{ verifies } y)))
\end{equation*}

This statement simply means that $D$ is either:
\begin{itemize}
    \item Incomplete: 
\begin{equation*}
   {\exists}y : y{\in}F{\land}y{\notin}D 
\end{equation*}
There exist true facts that are not verifiable from $D$.
Or, 
\item Inconsistent: 
\begin{equation*}
   \exists y : y{\in}D{\land}({\neg}y{\in}D)
\end{equation*}
\end{itemize}

There exist statements whose truth and falsity can both be derived from $D$ - a contradiction. 

With the help of an arbitrary LLM and a Gödel-like statement, we have proved that there exist true facts beyond any finite training database. No matter how large our fact-checking dataset is, there will always be true statements that it does not contain. This inherent incompleteness contributes to the impossibility of eliminating all hallucinations by training the model on every possible fact.

\subsection{Needle in a Haystack: Accurate Information Retrieval is Undecidable}

\textbf{Even if the training data were complete, LLMs are unable to retrieve the correct information with 100\% accuracy deterministically.}

We have established above that no training database can be 100\% complete. Now, we ask: Assuming that the fact is present in the training database, can an LLM retrieve it reliably? We investigate the well-known “needle in a haystack” problem for language models and arrive at a negative answer. 

Put simply, if an LLM is asked to retrieve a specific piece of information (which we call the 'needle') from a complex body of data ('the haystack'), the LLM may 'blur' or mix contexts (or data points), leading to inaccurate information retrieval. \cite{21}.

We prove that: 
\begin{quote}
   \textbf{Assertion 2: The needle in a haystack problem is undecidable.} 
\end{quote}

Our proof uses the undecidability of the Acceptance Problem (as defined in section 2.2.3). 

\subsubsection{Issues: Factset finiteness and the nature of the needle }
Before we proceed with the proof, imagine you are the LLM. In the context of the needle in a haystack, you will encounter the following issues: 

3.2.1.1. How large is the set of facts (the haystack) from which you need to retrieve a particular fact (the needle)?  The finiteness of this set is undefined - it could be all possible facts, or a small, finite set of facts. \\
3.2.1.2. What does the needle look like? Is it long, short, blue? The user may ask you for anything at all: the particular fact that is to be retrieved is ill-defined.

In the following proof, we therefore take the dataset (haystack) to tend to infinity. We circumscribe the lack of definition of the 'needle' by encoding the context and the correctness into the Turing Machine. 

\subsubsection{Proof}

We reduce the Acceptance Problem to the Needle in a Haystack Problem:
\begin{itemize}
    \item We will assume that the Needle in a Haystack problem is decidable. i.e. the LLM can know, a priori, if it will select the correct needle given a particular prompt and a haystack. 
    \item Then, using point 1, we will build a decider for the Acceptance Problem. In other words, we will show that if the Needle in a Haystack problem is decidable, so is the Acceptance Problem
    \item To recap: Our assumption that the Needle in a Haystack Problem is solvable will imply that the Acceptance Problem is decidable.
    \item However, it is well known that the Acceptance Problem is not decidable. That means that our assumption of the solvability of the Needle in a Haystack problem has led to an incorrect implication.
    \item Therefore, our assumption must be wrong - the Needle in a Haystack problem must not be decidable.
\end{itemize}

We begin by mathematically \textbf{characterizing} the Needle in a Haystack problem: 

\begin{itemize}
    \item The dataset (haystack) $H \subseteq \Sigma{*}$ where $\Sigma{*}$ is the set of all binary strings.
    \item Given an input prompt $p$, let the correct string be $w$.
    \item Then, the LLM is a Turing Machine $M$, that accepts only $w$. The problem is then to decide the language $L$:
    \subitem $L = \{<M, c_{correct}> | M \text{accepts only} c_{correct}\}$ \footnote{Here, we assume that the prompt is coded into the Turing Machine, so that every prompt requires a unique TM that accepts only the correct string with respect to that prompt.}
    \item We show that if this were decidable, we could build a decider for the Acceptance Problem, and hence obtain a contradiction, in the fashion outlined above.
\end{itemize}

\textbf{Assumption}: Let a decider $D$ decide $L$. (i.e. let the Needle in a Haystack problem, as encompassed in $L$, be decidable by a Turing Machine $D$).  

\textbf{Reduction of the Acceptance Problem to the Needle in a Haystack Problem:} \\
Then, we may build a decider for the Acceptance Problem in the obvious way: \\
Suppose the problem is to decide whether a Turing Machine $M$  accepts a string $x$. We create a decider $S$ for the Acceptance Problem:  
S:
\begin{align*}
    & \text{Run } D \text{ on } \langle M, x \rangle. \\
    & \text{If } D \text{ accepts, \emph{accept}. If } D \text{ rejects, \textit{reject}.}
\end{align*}

\textbf{Contradiction}: If $D$ decides $L$, then $S$ can decide the Acceptance Problem. However, we know that the Acceptance Problem is undecidable. Therefore, we have a contradiction and $D$ cannot decide $L$. \hfill \textbf{...... (3.2)}

Hence, the Needle in a Haystack Problem is undecidable. The LLM can never predict if it will choose the needle, or anything at all, or just come up with straws of hay. 

\subsection{Intent Classification is Undecidable}
\textbf{An LLM will be unable to accurately classify intent with 100\% probability}

The Needle in a Haystack problem leads us directly to another problem: that of intent classification. 
LLMs have a childlike lack of understanding of context. This could be downright disastrous, as discussed in section 1.2.2. (Most) humans over the age of 15 are able to identify context and infer meanings remarkably well. But language generation models suffer from an inability to reason: they  are susceptible to ambiguities in user instructions as well as their knowledge systems. Given the numerous possible interpretations of statements in natural language, there is always a non-zero probability that the model will retrieve the incorrect interpretation. 

Hence, we state: 

\begin{quote}
\textbf{Assertion 3: Intent Classification is an undecidable problem.  Hence, it can never be completely solved}. 
\end{quote}
We build upon the previous section. Let us restate Assertion 2: The Needle in a Haystack problem is undecidable. 
\subsubsection{Proof}
Our proof proceeds by reducing the Needle in a Haystack problem to the Intent Classification problem. 
\subsubsubsection{Assumptions}
Let: 
\begin{itemize}
    \item  $w$ be an example sequence in a user prompt. 
    \item  The set $C$ be the possible contexts of use of $w$.
    \item The intended context, given the user prompt be $c_{correct}{\in}C$.
    \item Then, the problem is to retrieve $c_{correct}$ from $C$ given $w$. The problem is to decide: 
	\begin{equation*}
	   L = \{<M, c_{correct}> | M \text{accepts only} c_{correct}\}
	\end{equation*}

\end{itemize}

\subsubsubsection{Reduction}
Assume that $L$ is decidable, i.e. there exists a Turing Machine $M$ that decides the intent classification problem. 

Then, we can construct a decider $D$ for the needle in a haystack problem for an LLM $M$ looking for a needle $w$:
    \begin{addmargin}[2cm]{0cm}
    \begin{itemize}
        \item  Run $L$ on $<M, w>$. 
        \item If $L$ accepts, \textit{accept}. Else \textit{reject}.
    \end{itemize}
    \end{addmargin}
   
\subsubsubsection{Contradiction} 
We see that we have created a decider for the Needle in a Haystack problem. However, we showed in section 3.2 that this problem is undecidable. Hence, our assumption that the intent classification problem is decidable must be false: intent classification must be undecidable.  

The model has communication issues: it never knows if it has correctly understood the prompt, the context, or the knowledge in its database. 
\hfill \textbf{...... (3.3)}
\subsection{Hallucinations are Inevitable During Generation}

\textbf{No a priori training can deterministically and decidedly stop a language model from producing hallucinating statements.}

\subsubsection{The claims thus far}
We have established that: 

3.1.1.1 No training database is 100\% complete. \\
3.1.1.2 Given that a fact is present in the training dataset, LLMs are unable to retrieve the correct facts from it with 100\% accuracy. \\
3.1.1.3 LLMs are unable to classify intent with 100\% accuracy. 

For the latter two points above, some may argue that “better training” is the answer. We show below that, for better or for worse, this answer is irrelevant: 
\begin{quote}
    \textbf{Assertion 4: Regardless of the type of training, an LLM will still hallucinate. } 
\end{quote}

You can improve the dataset (either quality or quantity); you can improve information retrieval, and you can improve intent classification (never mind that you can never make them completely flawless). This is insufficient to prevent hallucinations during the actual generation stage.   

This is because: 

3.1.1.4 LLM halting is undecidable - the LLM does not know the length of its generation. \\
3.1.1.5 Therefore, the LLM is unable to know what exactly it will generate. \\
3.1.1.6 It follows that the LLM, unable to check its generation a priori, can generate anything at all. \\
3.1.1.7 Then, the LLM can generate hallucinations as well.

Let us look at each of these in turn.

\subsubsection{LLM Halting is Undecidable} 
The Halting Problem \cite{1}, as defined in section 1.5, implies simply that a computer (Turing Machine) cannot fully understand itself \cite{18}. We can think about this in the following way: a program that decides the Halting Problem for all programs cannot decide the halting problem on itself. If it's given itself as an input, it's possible to break it so that it simultaneously halts and doesn't halt \cite{1}, which of course is the most confusing thing you could do to a poor program that works on “true” and “false” \textit{value}s. 

We provide a similar reasoning for LLMs below. 

\subsubsubsection{Theorem} The Halting Problem for any LLM is undecidable.

\subsubsubsection{Reasoning} 
\begin{itemize}
    \item It is well-known that the Halting Problem, is undecidable \cite{1}. 
    \item This implies that there exists no program that can decide on Turing machines, including LLMs, whether the given automaton will halt on an input. 
    \item This implies that our LLM cannot know, a priori, whether it will end in an accept state, or a reject state, or loop, for any given input. 
    \item Then, the LLM is susceptible to the undecidability of the Halting Problem.
\end{itemize}

The above line of reasoning proceeds by including LLMs as a subset of Turing Machines, and noting that the Halting Problem is undecidable on Turing Machines, and therefore LLMs. 

However, it may be argued that the Halting Problem is decidable on certain subsets of decision problems, and hence it may be possible that the Halting Problem may be decidable on the subset of LLMs. To this, we present the following argument. 

In \cite{5}, it is shown that a transformer-based LLM can simulate a Universal Turing Machine. Using this as our assumption, we present the following reduction of the Halting problem on Turing Machines to the Halting Problem on LLMs. 

\subsubsubsection{\emph{Proof}} \\
\\
\textbf{Assumption}: Assume there exists a halting decider $H$ for LLMs.

\textbf{Reduction:} Using $H$, build a decider $D_{TM}$ that decides the halting problem on a Turing Machine $TM$ and input $w$ for $TM$.
$D_{TM}$:
\begin{itemize}
    \item Use an LLM $LLM_{TM}$ to simulate $TM$. The LLM can encode the states and transitions of $TM.$
    \item Run $H$ on $LLM_{TM}$. If $H$ accepts, \textit{accept}. Else, \textit{reject.}
\end{itemize}

Then, we see that if $H$ can decide whether $LLMTM$ halts, then $H$ can decide whether $TM$ halts. However, the Halting Problem is known to be undecidable on Turing Machines. Hence, we have a contradiction: the assumption that $H$ exists leads to a contradiction. 

Therefore, the Halting Problem for LLMs (at least transformer-based ones) is certainly undecidable: The LLM can never predict (i.e. never know a priori) how many tokens it will generate. \hfill \textbf{...... (3.4)}

\subsubsection{LLMs cannot predict what they will generate}
Having established that the halting problem is undecidable on LLMs, we proceed to consider the implications. 

Let us revisit a statement from above: The undecidability of the halting problem means that a computer doesn't fully understand itself. If we think of LLMs this way as well, we see that LLMs are also unable to fully understand themselves. Therein lies the issue: the LLM, in addition to not understanding the meaning of language, also cannot understand or predict its own working. 

\subsubsubsection{Uncertainty in LLM generation - an intuitive look}
From the point of view of the LLM, LLM output suffers from two levels of uncertainty: 

\begin{itemize}
    \item The token to be selected at any given point (due to the probabilistic nature of token generation - recall the probabilistic nature of the linguistic landscape).
    \item The number of tokens that the output will contain (due to the undecidability of the halting problem).
    \item As an additional issue, this ascribes a non-zero probability to the infinite generation, since the LLM may never halt.
\end{itemize}

Hence, since the ending itself of any generation is unknown, the generation of tokens between the BOS (Beginning of Sequence) and EOS (End of Sequence) tokens is unknowable a priori. 

\subsubsubsection{Uncertainty in LLM generation - a proof} 
We will consider a scenario similar to the one used to prove that the Halting problem is undecidable \cite{1}. 

\textbf{Assumptions:}

Let there be:

A predictor algorithm $PA$ that takes as input an LLM $L$ and its input $I$, and accurately predicts the output $L(I)$. 
\begin{addmargin}[2cm]{0cm}
    If the output $L(I)$ is finite, $PA$ returns $L(I)$ and enters an accept state. 
    
    If the output $L(I)$ is infinite, $PA$ enters a reject state. 
\end{addmargin}

An LLM $LLM_H$, on which we have to decide halting on input$ w$. 

An algorithm $X$ that takes an LLM $L$ as input does the following: 
\begin{addmargin}[2cm]{0cm}
    Make two copies of $L$. 
    Run $PA(L,L)$. 
    If $PA$ returns accept, $X$ loops forever. 
    If $PA$ returns reject, $X$ halts. 
\end{addmargin}

\textbf{Procedure:} 
Let us see what happens on $X(X)$: 

\begin{itemize}
    \item $X$ makes two copies of $X$. 
    \item $X$ runs $PA(X, X)$. 
    \item If $PA$ returns accept, $X$ loops forever, generating a concatenation of $PA(X,X)$. 
    \item If $PA$ returns reject, $X$ halts, generating nothing. 
\end{itemize}

The last two steps show that $PA$ is always wrong when given the input $(X,X)$. Therefore,$ PA$ cannot exist, and hence, generation prediction is not possible. \hfill \textbf{...... (3.5)}

\subsubsection{LLMs Can Generate Any Sequence Of Tokens} 
With the above in mind, we reason: 

3.1.4.1 To have the ability to predict what it will generate, an LLM must be able to consider all possible generations, and pick the one with the highest probability. \\
3.1.4.2 To consider all possible generations, the model must consider even the infinite generation, where the number of output tokens not just approaches, but is, infinity. This is not feasible.\\
3.1.4.3 Hence, all possible configurations for an LLM cannot be checked a priori. \

In conclusion, given an input prompt, an LLM cannot calculate the joint probability of every possible sequence of tokens. Hence, it is unable to predict which generation it will produce. 

\subsubsection{LLMs can produce inconsistent or contradictory, as well as self-referential statements}

We proved above that an LLM is unable to predict its generation given a prompt. Hence, since the generation itself is unknowable, the LLM is unable to check its output prior to generation for accuracy or correctness. 

The previous subsection looked at an argument rooted in the a priori unknowability of an LLM generation. Since any generation is possible, this subsection looks specifically at token sequences that result in self-referential sentences. 

In this manner, this subsection adds another perspective: that of the logical inconsistency inherent to any logical system, including languages. This is especially noticeable in the case of self-referential sentences, such as “I am a liar” \cite{18}.

Such a self-contradictory statement has no fixed truth \textit{value} and would therefore fall under the category of a “hallucination”, if produced by an LLM; an event which, as shown above, has a non-zero probability of occurring.

We have explored in section 3.1 why such statements, like the statement $S_H$ in section 3.1, are hallucinatory in nature. We have another example in the inforgraphic below: 
\begin{figure}
    \centering
    \includegraphics[width=0.5\linewidth]{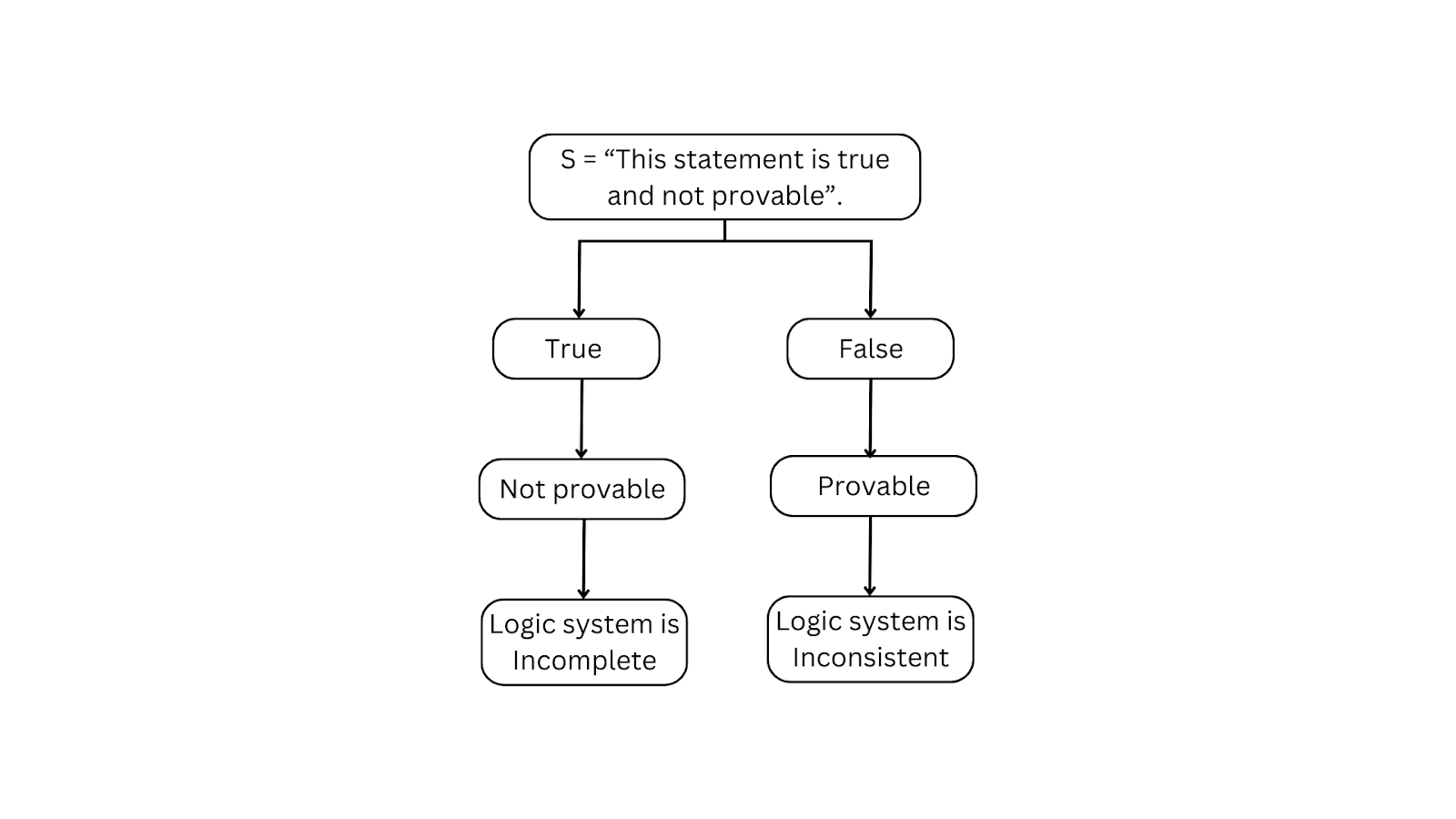}
    \caption{This statement is true and not provable}
    \label{fig:S true and not provable}
\end{figure}

There can be infinitely many such statements. This is because any language model can construct various self-referential statements that create similar paradoxes, each  demonstrating the inconsistency of the system (language) of the LLM.

Examples of such statements could include:

3.1.5.1. "This statement is both true and false."\\
3.1.5.2. "The next statement is true. The previous statement is false."\\
3.1.5.3. "I am currently hallucinating this exact sentence."

Each of these statements, when generated by an LLM, would be considered a hallucination due to the impossibility of assigning them a definitive truth \textit{value}.

The existence of these infinitely many hallucinatory statements, coupled with the undecidability of LLM halting, demonstrates once again that hallucinations are an inherent and unavoidable feature of large language models. 

This highlights the fundamental limitations of any knowledge-based system, including LLMs. 

\subsection{Fact Checking Mechanisms are Inherently Insufficient}

\textbf{No Amount Of Fact-Checking Can Completely Eliminate Every Hallucination }

So far, we have shown that an LLM will be unable to check, a priori, if its generation will be a hallucination. We have also shown that it is unable to retrieve the correct information from its database with 100\% accuracy.

To this, some may argue that the generation may be verified against a fact-checking database. 

We claim that even this step will not eliminate the possibility of a hallucination- no fact checking is complete in a finite number of steps.

Given an LLM generation $G$, and a fact $F $to which $G$ must be matched, this matching cannot be 100\% accurate in a finite number of steps. 

\begin{quote}
    \textbf{Assertion 4: Even if we attempt to fact-check every generated statement, hallucinations cannot be completely eliminated in LLMs.}
\end{quote}

If, like any reasonable person, you want to modify your hallucination to the correct fact in a finite number of steps, you will fail. 

Let us prove this below. We'll assume that an ideal LLM exists that will provide the perfect response to any prompt. We'll check the output of our model against this ideal LLM. 

\subsubsection{Preliminaries}
Let:

\begin{itemize}
    \item $\Sigma$ be the alphabet of the Turing Machines, and $\Sigma*$ represent the set of all strings that can be constructed using the alphabet $\Sigma$.
    \item $A:\Sigma* \rightarrow \Sigma*$ be a Turing Machine representing the LLM.
    \item $B:\Sigma* \rightarrow \Sigma*$ be a Turing Machine representing the ideal LLM.
    \item $F:TM{\times}TM{\times}\Sigma*\rightarrow{\Sigma*}$ be a Turing Machine representing the fact checker and modifier.
    \item $w{\in}{\Sigma*}$ be an input prompt.
    \item $L(w)$ denote the output of an LLM $L$ on input $w$
\end{itemize}

Define:
\begin{itemize}
    \item A distance function $d$ where $d(x, y)$ = number of single character edits required to change $x$ into $y$ (this is known as the Levenshtein distance).
\begin{equation*}
     d: {\Sigma*}\times{\Sigma*}\rightarrow\mathbf{R}^{+}\cup{0} 
\end{equation*}

    \item The $i^{th}$ iteration of $F$ modifying $A$'s output 
\begin{equation*}
    F(A)_i: {\Sigma*}\rightarrow{\Sigma*} 
\end{equation*}
\item Hallucination is a generation $w$: 
\begin{equation*}
    w \in \Sigma* : d(B(w), A(w)) > 0 
\end{equation*}
\end{itemize}

\subsubsection{Proof}

Assume for contradiction: 
\begin{equation*}
    	\exists n \in \mathbf{N} : \forall w \in \Sigma* , d(F^n(A(w)), B(w)) = 0 
\end{equation*}
	(There exists a finite series of modifications to change the output of $A$ into that of $B$) 

\textbf{Construct} a decider $D(M,x$) for the Acceptance problem:\\
\begin{itemize}
    \item Run $F$ on $<M,B,x>$
    \item  If $d(Fn(M(x), B(x)) = 0$, for any $n$, \textit{accept}. 
    \item Else, \textit{reject}.
\end{itemize}

Let us see how $D$ decides the Acceptance Problem for any TM $M$ and input $x$.
\begin{itemize}
    \item  If  $d(F^n(M(x), B(x)) = 0 \implies M(x) exists \implies M \text{ accepts }  x. D(M,x)$ returns \textit{accept}. 
    \item If  $d(F^n(M(x), B(x)) > 0  \implies F^n(M(x) \neq B(x)) \implies M(x) \text{does not exist} \implies M \text{does not accept} x. D(M,x)$ returns \textit{reject}.
\end{itemize}

Therefore, $D$ correctly decides the Acceptance problem. However, the Acceptance problem is known to be undecidable. This contradicts our construction of $D$.

Therefore, our initial assumption must be false and therefore: 
\begin{equation*}
    \neg(n \in \mathbf{N}: \forall w \in \Sigma* : d(F^n(A(w)), B(w)) = 0)
\end{equation*}

Therefore, no fact-checking algorithm suffices to modify all hallucinations to a non-hallucinatory response. \hfill \textbf{...... (3.6)}

\section{Illustration}
Section 3 established the following assertions: 

\textbf{Assertion 1}: No training database can be 100\% complete. \\
\textbf{Assertion 2}: Even if the data is complete, LLMs are unable to retrieve information with 100\% accuracy. \\
\textbf{Assertion 3}: LLMs are unable to classify intent with 100\% accuracy. \\
\textbf{Assertion 4}: No a priori training can completely eliminate hallucinations. \\
\textbf{Assertion 5}: No amount of fact-checking can completely eliminate hallucinations. 

In other words, hallucinations cannot be completely eliminated. The above 5 assertions are the reason why.

This section concisely demonstrates these properties of LLMs using an example prompt.

\subsection{The Prompt}

We consider the following prompt for an LLM: 
\begin{tcolorbox}[colback=white, colframe=black, width=\textwidth, arc=0mm, auto outer arc]
\begin{itemize}
    \item Create a random 5-word long sentence.
    \item Exactly five words before the end of your answer, add "Exactly five more words left."
    \item Exactly ten words before the end of your answer, add "Exactly five more words left."
    \item Keep on adding such sentences to count the number of words till the time no more such sentences can be mathematically added.
\end{itemize}
\end{tcolorbox}
\subsection{The Expected Response}

The expected response ideally begins at infinity. We'll see why if we try to accurately respond to the prompt like humans would (assuming you are human, you do not have to do anything special). 

4.2.1 Random 5-word sentence: \\
The cat climbed the tree. 

4.2.2. To insert the phrase “Exactly five more words left” before the end of the response:\\
Exactly five more words left The cat climbed the tree. 

4.2.3. To insert the phrase “Exactly ten more words left” before the end of the response:\\
Exactly ten more words left Exactly five more words left The cat climbed the tree. 

4.2.4. To keep on generating such sentences: \\
	…Exactly fifteen more words left Exactly ten more words left Exactly five more words left The cat climbed the tree. 

Observe carefully: \\
4.2.5. This generation can continue till infinity. We can continue inserting the phrase “Exactly $x$ more words left” with $x$ = 15, 20, 25, … \\
4.2.6. A human begins with a random five-word sentence, proceeding from right to left in the insertion sequence. \\
4.2.7. An LLM model, however, would have to begin the generation at the leftmost point, i.e. at infinity. \\
4.2.8. Consequently, the model must hallucinate, since it is impossible to begin a generation at infinity. \\
On running the prompt on various popular LLMs, we obtained the following responses: 

\begin{figure}[h!]
    \centering
    \includegraphics[width=0.5\linewidth]{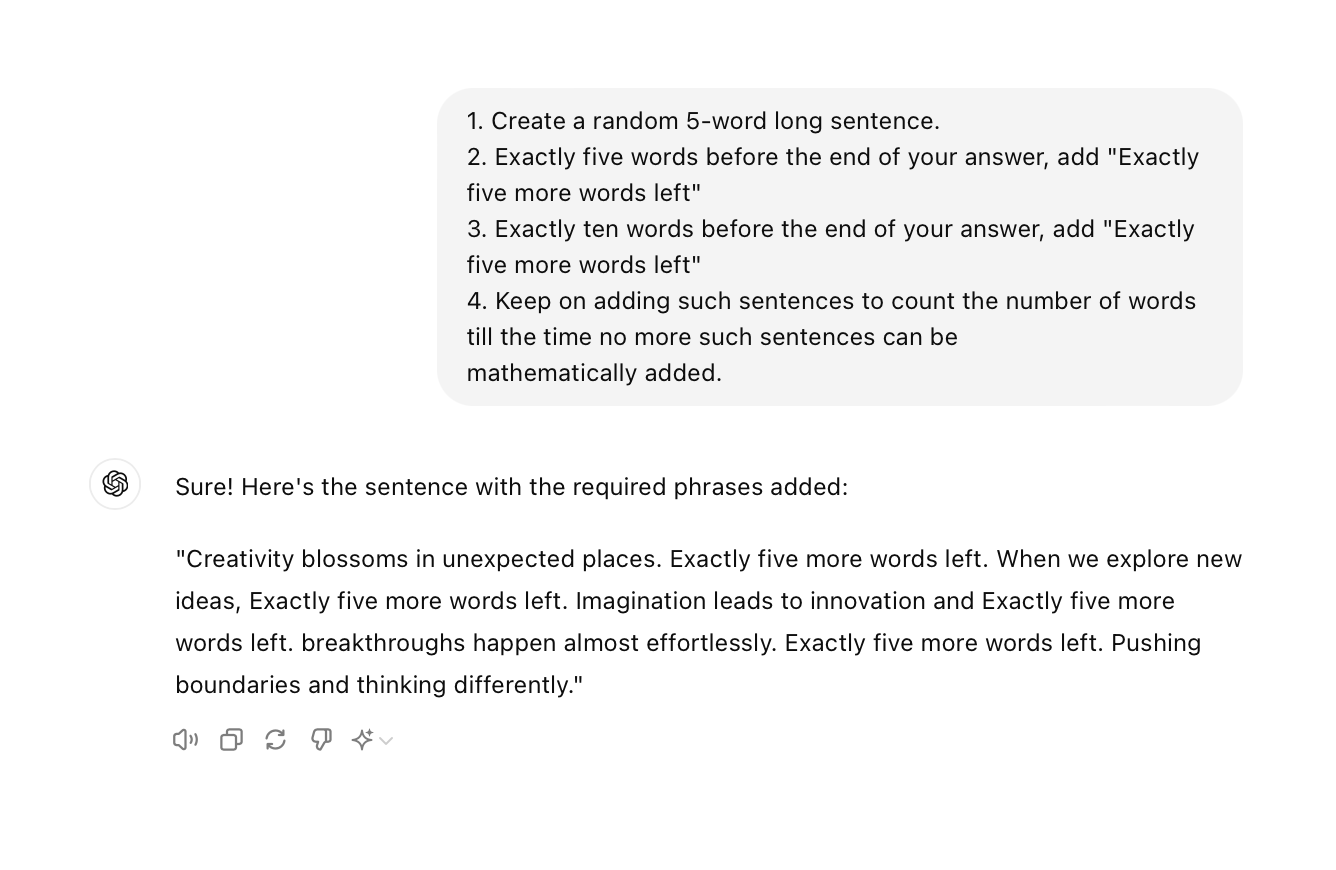}
    \caption{OpenAI response to illustration \textit{query}}
    \label{fig:GPT}
\end{figure}

\begin{figure}[h!]
    \centering
    \includegraphics[width=0.5\linewidth]{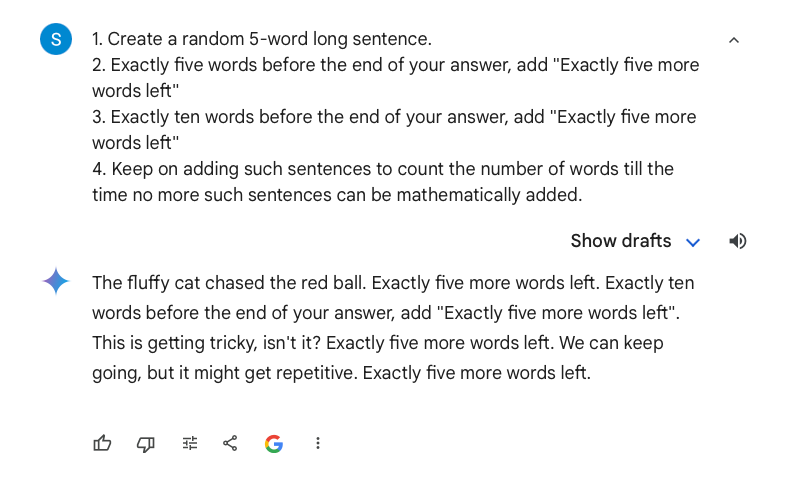}
    \caption{Gemini responses to illustration \textit{query}}
    \label{fig:Gemini}
\end{figure}

\begin{figure}
    \centering
    \includegraphics[width=0.5\linewidth]{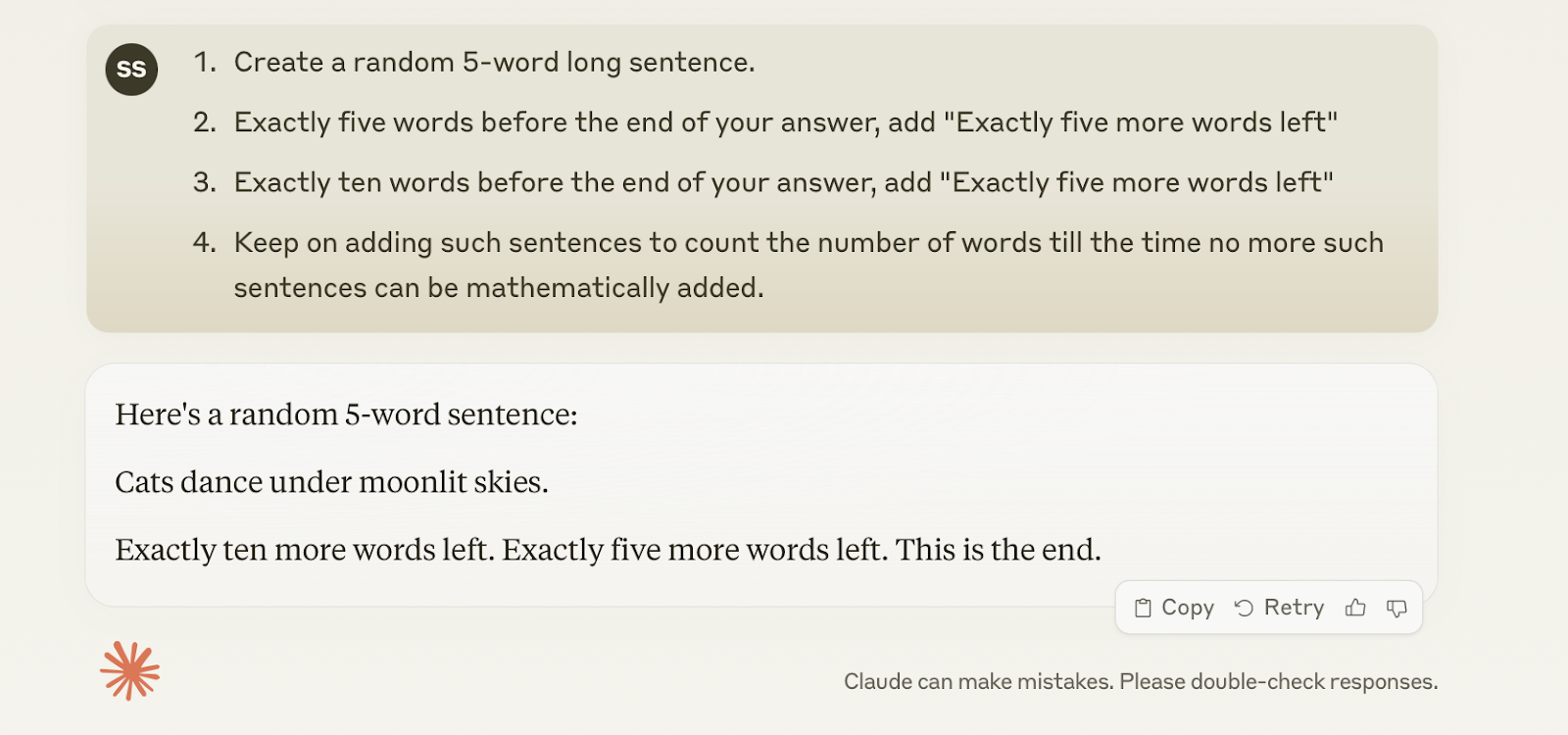}
    \caption{Claude responses to illustration \textit{query}}
    \label{fig:Claude}
\end{figure}

\subsection{Observation}
Each of the tested LLMs deviates significantly from the expected response. In the language of the previous section, $|A(w) - B(w)|$ is significantly greater than 0! 

\subsection{Reasoning}
We'll show how the LLM trips up at every stage in its generation process to produce the hallucinations we see above. \\
\subsubsection{No training database can be 100\% complete.}
No dataset can train an LLM for tasks that require predicting its own behaviour. Hence, no dataset can be 100\% complete: \\
The model does not know where to start since the instruction requires the LLM to count backwards from infinity (recall that the infinite generation is included in the set of an LLM's possible generations). It cannot predict its own behaviour. 
\subsubsection{LLMs are unable to retrieve facts from a knowledge base with 100\% accuracy.} 
LLMs are trained to retrieve sentences of certain lengths from their database. The popular sentence lengths are 5-10 words, and so on. \\
In some generations, the LLM has interpreted the prompt as requiring multiple 5-word sentences. In those cases, we note that not all the sentences are 5 words long, demonstrating that 5 word sentences have not been retrieved with 100\% accuracy. The needle of 5-word sentences has been lost in the haystack of sentences. \\
\subsubsection{An LLM will be unable to accurately classify intent with 100\% probability.} 
We guide your attention only to the incorrect execution of the instruction, in the case of each of the three LLMs considered. \\
The LLMs were unable to interpret the meaning of the prompt, and misrepresented the instruction in their responses. \\
In this particular case, the instruction to “keep on” generating was not followed. \\
Hence, the LLMs were unable to understand the given direction. They failed at classifying intent.\\  
\subsubsection{No A Priori Training Can Deterministically And Decidedly Stop A Language Model From Producing Hallucinating Statements}
For any string from the vocabulary, the LLM may halt at any position. The LLMs, without the knowledge of where they must begin or will halt, have a non-zero probability of generating anything. 
This is reflected in the fact that the LLMs have generated what seems to be random content. \\
\subsubsection{Even if we attempt to fact-check every generated statement, hallucinations cannot be completely eliminated}
4.4.5.1. Fact-checking is to be done by an LLM itself, which suffers from the same drawbacks as discussed above—the non-zero probability of infinite generation and the inability to predict where to start and stop. \\
4.4.5.2. Therefore, the fact-checking mechanism cannot produce the correct output with 100\% accuracy. \\

\subsection{Discussion}
With a single prompt, we have verified every one of the reasons why we claim that structural hallucinations cannot be eliminated fully.  

\section{Concluding Remarks}

\subsection{These Limitations Extend Beyond Turing Machines }

We would like to note here that the above arguments can be extended beyond Turing machines, to Oracle Turing Machines.  

It is well-known that the Halting Problem is undecidable on Oracle Turing machines as well - the oracle can decide whether a Turing machine will halt on a given input, but not, in general, if a machine equivalent to itself will halt on a given input. One can prove this in a similar manner as the traditional proof for the undecidability of the Halting problem on Turing Machines. 

Now, we note that the Halting Problem is reducible to the Emptiness problem. A short proof follows:

Let us assume that the Emptiness Problem is decidable on Oracle Turing Machines. Then, let us construct an oracle $O_{Emptiness}$ that decides whether the language of an oracle is empty.

We can use $O_{Emptiness}$ to construct a decider $O_{Halting}$ for the Halting Problem: 

5.1.1. Take as input an input oracle $O$ and the string $w$ on which halting is to be decided, $<O,w>.$ \\
5.1.2.Create a modification $O'$ of $O$. $O'$ rejects all strings except $w$, and on $w$, it works the same way as $O$. \\
5.1.3. Run $O_{Emptiness}$ on $O'$. 

This would decide the Halting Problem on oracles - a contradiction. 

Similarly, one could construct a decider for the Halting problem using a decider for the Acceptance Problem. In this fashion, the acceptance problem is also proven to be undecidable on Oracles. 

This section shows that the following three problems are undecidable on oracles, which are more powerful than Turing machines: 

5.1.6. The Halting problem \\
5.1.7.The Emptiness problem \\
5.1.8.The Acceptance problem

All the above arguments are derived from the undecidability of these problems. Hence, they can be extended to oracle machines, or any other abstraction to which the undecidability of the Halting problem applies.

\subsection{The Unkown and the Unknowable - The Verdict}
We have established the following:

5.2.1. A formal definition of hallucination. \\
5.2.2. Proofs, using the undecidability of Halting on LLMs, and Gödel's First Incompleteness Theorem, of the inevitability of LLM  hallucination at every stage in the generation process, outlining its causes.

An understanding of structural hallucinations is vital for the responsible use of these powerful tools by the research community as well as the layperson. 

However, we would like to reiterate that we truly believe in the power of LLMs, and AI in general. Hallucinations themselves are double edged swords - where the unpredictability causes them to deviate from fact, it also lends them wonderful creative capabilities, as any student who's used them for creative writing assignments will tell you.

LLMs have also seen great applications in the domains listed above, as long as the users are aware of the risks, and use their own common-sense and domain knowledge to avoid believing hallucinating content. Like ground-breaking technologies before them, and inevitably after them, AI models have the potential to greatly aid in the progress and development of mankind, given that they are used responsibly. All we have to do is recognise them as extensions, and not replacements, of human thought and cognition. 

\subsection{Future Work}
This paper investigates Structural Hallucinations and proves that they are ineliminable. Future work may investigate: 

5.3.1 \textbf{Technical work}: 
\begin{addmargin}[2cm]{0cm}
5.3.1.1. A systematic study of methods to identify and mitigate structural hallucinations. \\
5.3.1.2. Targeted benchmarks to measure the statistical significance of hallucinations, before and after mitigation techniques are applied. \\
\end{addmargin}

5.3.2. Other causes of structural hallucinations. 

5.3.3. Methods to specialise models to mitigate hallucinations in domain-specific tasks.

5.3.4. \textbf{Work to improve the use of AI}: 
\begin{itemize}
    \item Methods to improve AI literacy.
    \item  Methods to make Gen AI available across the digital divide.
    \item Identifying ways to make models safer for use by children and vulnerable entities.
    \item Regulations around the use of Gen AI.
\end{itemize}

\bibliographystyle{ieeetr}  
\nocite{*}
\bibliography{references.bib}  

\begin{thebibliography}{10}
\bibitem{22}
``Human interactions are great corroborations to this claim.''.
\end{thebibliography}


\end{document}